\documentclass[10pt]{article} 
\usepackage{hyperref}
\usepackage{url}
\usepackage{graphicx}
\usepackage{subfigure} 
\usepackage{amsmath,amssymb,amsthm,morefloats,enumerate}

\newcommand{\alg}{\mathbb A}
\newcommand*{\LargerCdot}{\raisebox{-0.25ex}{\scalebox{1.2}{$\cdot$}}}

\newtheorem{definition}{Definition}[section]

\title{Universal halting times in optimization and machine learning}

\author{
\begin{minipage}{.3\textwidth}\centering
Levent Sagun \\
Mathematics Department\\
New York University \\
\texttt{sagun@cims.nyu.edu}
\end{minipage}
\begin{minipage}{.34\textwidth}\centering
Thomas Trogdon \\
Department of Mathematics\\
University of California, Irvine\\
\texttt{ttrogdon@math.uci.edu} 
\end{minipage}
\begin{minipage}{.35\textwidth}\centering
Yann LeCun \\
Computer Science Department \\
New York University \\
\texttt{yann@cs.nyu.edu}
\end{minipage}
}

\begin{document}

\maketitle

\begin{abstract} 

The authors present empirical distributions for the halting time (measured by the number of iterations to reach a given accuracy) of optimization algorithms applied to two random systems: spin glasses and deep learning. Given an algorithm, which we take to be both the optimization routine and the form of the random landscape, the fluctuations of the halting time follow a distribution that, after centering and scaling, remains unchanged even when the distribution on the landscape is changed. We observe two qualitative classes: A Gumbel-like distribution that appears in Google searches, human decision times, the QR eigenvalue algorithm and spin glasses, and a Gaussian-like distribution that appears in conjugate gradient method, deep network with MNIST input data and deep network with random input data. This empirical evidence suggests presence of a class of distributions for which the halting time is independent of the underlying distribution under some conditions.

\end{abstract} 

\section{Introduction}
\label{intro}

In this letter we discuss the presence of universality in optimization algorithms. More precisely, we analyze the number of iterations of a given algorithm to optimize (or approximately optimize) an energy functional when the functional itself and the initial guess are random. We consider the following iterative routines: conjugate gradient for solving a linear system, gradient descent for spin glasses, and stochastic gradient descent for deep learning. 

A bounded, piecewise differentiable random field (See \cite{adler2009random} for an account on the connection of random fields and geometry), where the randomness is non-degenerate, yields a landscape with many saddle points and local minima.  We refer to the value of the landscape at a given point as the \emph{energy}. Given a moving particle on the landscape that takes a sequence of steps to reach a (local) minimum, an essential quantity is the time (or number of steps) the particle requires until it stops. We call this the \textit{halting time}. Many useful bounds on the halting time are known for convex cases, where the stopping condition produces a halting time that is, essentially, the time to find the minimum. In non-convex cases, however, the particle knows only the information that can be calculated locally. And a locally computable stopping condition, such as the norm of the gradient at the present point, or the difference in altitude with respect to the previous step, can lead the algorithm to locate a local minimum. This feature allows the halting time to be calculated in a broad range of non-convex, high-dimensional problems, even though the global minimum may not be located. 

A prototypical example of such a random field is the class of polynomials with random coefficients. Spin glasses and deep learning cost functions are special cases of such fields that yield very different landscapes. We emphasize that polynomials with random coefficients are not only a broad class of functions, but even they are hard to study mathematically in any generality. Therefore, in order to capture essential features of such problems, we focus on their subclasses that are well studied (spin glasses) and practically relevant (deep learning cost functions).

The halting time in such landscapes, when normalized to mean zero and variance one (subtracting the mean and dividing by the standard deviation), appears to follow a distribution that is independent of the random input data --- the fluctuations are universal. In statistical mechanics, the term ``universality'' is used to refer to a class of systems which, on a certain macroscopic scale, behave statistically the same while having different statistics on a microscopic scale. An example of such a law is the central limit theorem (CLT), which states that the sums of observations tend to follow the same distribution independent of the distribution of the individual observations, as long as contribution from individual observations is reasonably small. It may fail to hold, if the microscopic behavior is not independent, does not have a finite second-moment, or if we move beyond summation.

\subsection{Results}

The focus of this work is an attempt to put forward additional cases where we \textit{see} universality with an emphasis on routines and landscapes from machine learning. We present concrete evidence that the halting time in such optimization problems is universal (Sections~\ref{s:spin} and \ref{s:mnist}). But, in the spirit of the potential failure of the CLT in degenerate cases, we show a degenerate case for the conjugate gradient algorithm (Section~\ref{s:cg}) in which halting time fails to follow a universal law.

\begin{figure}[tbp]
\begin{center}
\includegraphics[width=0.6\linewidth]{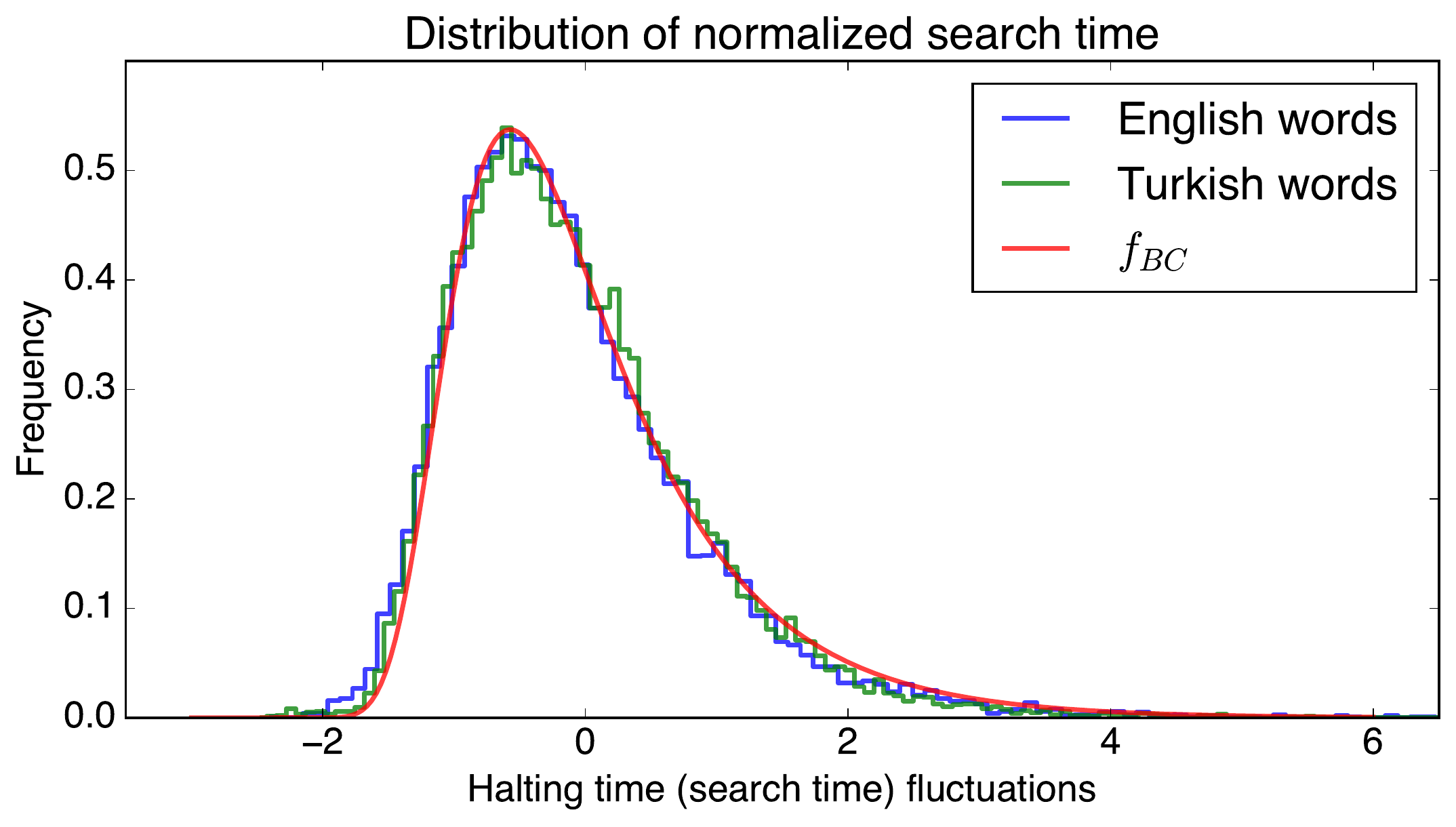}
\end{center}
\caption{\label{f:google} Search times of randomly selected words from two ensembles is compared with the curve $f_{\mathrm{BC}}$ in \cite{bakhtin2012neural} that is estimated from the decision times in an experiment conducted on humans. It is evident that more observations have yet to be made in identifying the underlying principles of the algorithms that are increasingly part of our life.}
\end{figure}

Another example of halting time universality is in the cases of observed human decision times and {\tt Google}\textsuperscript{TM} query times. In \cite{bakhtin2012neural} the time it takes a person make a decision in the presence of visual stimulus is shown to have universal fluctuations. The theoretically predicted distribution $f_{\mathrm{BC}}$ for this experiment is a Gumbel distribution.  In a surprising connection, we randomly sampled words from two different dictionaries and submitted search queries to the {\tt Google}\textsuperscript{TM} search engine. The times it takes {\tt Google}\textsuperscript{TM} to present the results are recorded. The normalized search times closely follow the same Gumbel curve, after normalizing to mean zero and variance one, see Figure~\ref{f:google}.


In the cases we observe below, we find two qualitative universality classes: (1) A non-symmetric Gumbel-like distribution that appears in Google searches, human decision times and spin glasses, and (2) a symmetric Gaussian-like distribution that appears in the conjugate gradient algorithm and in deep learning. 


\subsection{Definition of universality}

\begin{definition}
An algorithm $\alg$ consists of both a random cost function $F(\mathbf x, w)$ where $\mathbf x$ is a given random input and an optimization routine that seeks to minimize $F$ with respect to $w$.
\end{definition}

To each algorithm we attach a precise $\epsilon$-dependent halting criteria for the algorithm. The halting time, which is a random variable, is the time (\emph{i.e.} the number of iterations) it takes to meet this criteria. Within each algorithm there must be an intrinsic notion of dimension which we denote by $N$. The halting time $T_{\epsilon,N,\alg,E}$ depends on $\epsilon$, $N$, the choice of algorithm $\alg$, and  the ensemble $E$ (or probability distribution). We use the empirical distribution of $T_{\epsilon,N,\alg,E}$ to provide heuristics for understanding the qualitative performance of the algorithms.

The presence of universality in an algorithm is the observation that for sufficiently large $N$ and $\epsilon = \epsilon(N)$, the halting time random variable satisfies
\begin{align}\label{univ}
  \tau_{\epsilon,N,\alg,E}:=\frac{T_{\epsilon,N,\alg,E}-\mathbb E[T_{\epsilon,N,\alg,E}]}{\sqrt{\mathrm{Var}(T_{\epsilon,N,\alg,E})}} \approx \tau^*_{\alg},
\end{align}
where $\tau^*_{\alg}$ is a continuous random variable that depends only on the algorithm. The random variable $\tau_{\epsilon,N,\alg,E}$ is referred to as the \emph{fluctuations} and when such an approximation appears to be valid we say that $N$ and $\epsilon$ (and any other external parameters) are in the \emph{scaling region}. We note that universality in this sense can also be used as a measure of stability in an algorithm. Some remarks must be made:



\begin{itemize}
\item A statement like \eqref{univ} is known to hold rigorously for a few algorithms (see \cite{Deift2016,deift2017universality}) but in practice, it is verified experimentally.  This was first done in \cite{DiagonalRMT} and expanded in \cite{deift2014universality} for a total of 8 different algorithms.
\item The random variable $\tau^*_{\alg}$ depends fundamentally on the functional form of $F$. And we only expect \eqref{univ} to hold for a restricted class of ensembles $E$.
\item $T_{\epsilon,N,\alg,E}$ is an integer-valued random variable. For it to become a continuous distribution limit must be taken. This is the only reason $N$ must be large --- in practice, the approximation in \eqref{univ} is seen even for small to moderate $N$.
\end{itemize}

\subsection{Core examples: Spin glass Hamiltonians and deep learning cost functions}

A natural class of random fields is the class of Gaussian random functions on a high-dimensional sphere, known as $p$-spin spherical spin glass models in the physics literature (in the Gaussian process literature they are known as isotropic models). From the point of view of optimization, minimizing the spin glass Hamiltonian is fruitful because a lot is known about its critical points. This allows one to experiment with questions regarding whether the local minima and saddle points, due to the non-convex nature of landscapes, present an obstacle in the training of a system. 


Following the asymptotic proof in \cite{auffinger2013random}, the local minima of the spin glass Hamiltonian lie roughly at the same energy level. Moreover, the values of the ground states and the exponential growth of the average of the number of critical points below any given energy level have been established. It turns out, when the dimension (or the number of spins) is large, the bulk of the local minima tend to have the same energy which is slightly above the global minimum. This level is called the \textit{floor} level of the function. An optimization simulation for this model can only locate the values at the floor level, and not deeper. This same phenomenon is present in the optimization of the MNIST\footnote{MNIST is a database of handwritten numerical digits that is commonly used as a means of benchmarking deep learning networks \cite{MNIST}.} classification problem \cite{sagun2014explorations}.   We emphasize that this striking similarity is only at the level of analogy, and the two systems are in fact vastly different. To the best of the authors' knowledge, there are no known theoretical arguments that connect spin glass Hamiltonians to deep learning. However, the feasibility of the observation of the floor level in the optimization of the two problems may give a hint at possible universal behaviors that can also be observed in other systems.



\begin{itemize}
\item Given data (i.e., from MNIST) and a measure $L(x^\ell,w)$ for determining the cost that is parametrized by $w\in \mathbb{R}^N$, the training procedure aims to find a point $w^*$ that minimizes the empirical training cost while keeping the test cost low. Here $x^{\ell} \in Z$ for $\ell \in  \{1,...,S\}$, where $Z$ is a random (ordered) sample of size $S$ from the training examples. Total training cost is given by
\begin{equation}
 F(Z,w) = \mathcal{L}_{\text{Train}}(w) = \frac{1}{S} \sum_{\ell=1}^S L(x^{\ell},w) \label{train}.
\end{equation} 
\item Given couplings $x_{(\LargerCdot) } \sim \text{ Gaussian}(0,1)$ that represent the strength of forces between triplets of spins. The state of the system is represented by $w\in S^{N-1}(\sqrt{N}) \subset \mathbb{R}^N$. The Hamiltonian (or energy) of the simplest complex\footnote{2-spin spherical spin glass, sum of $x_{ij}w_iw_j$ terms, has exactly $2N$ critical points. When $p\geq3$, $p-$spin model has exponentially many critical points with respect to $N$. For the latter case, complexity is a measure on the number of critical points in an exponential scale. Deep learning problems are suspected to be complex in this sense.} spherical spin glass model is given by: 
\begin{equation}
F(x_{(\LargerCdot)}, w) = H_{N}(w) = \frac{1}{N}\sum_{i, j, k}^Nx_{ijk}w_{i}w_{j}w_{k} \label{ham}.
\end{equation}
\end{itemize}

The two functions are indeed different in two major ways. First, the domain of the Hamiltonian is a compact space and the couplings are independent Gaussian random variables whereas the inputs for \eqref{train} are not independent and the cost function has a non-compact domain. Second, at a fixed point $w$, variance of the function $\mathcal{L}_{\text{Train}}(w)$ is inversely proportional to the number of samples, but the variance of $H_{N}(w)$ is $N$. As a result a randomly initialized Hamiltonian can take vastly different values, but randomly initialized costs tend to have very similar values. The Hamiltonian has macroscopic extensive quantities: its minimum scales with a negative constant multiple of $N$. In contrast, the minimum of the cost function is bounded from below by zero. All of this indicates that landscapes with different geometries (glass-like, funnel-like, or another geometry) might still lead to similar phenomena such as existence of the floor level, and the universal behavior of the halting time. 
 



\section{Empirical observation of universality}

We discuss the presence of universality in algorithms that are of a very different character. The conjugate gradient algorithm, discussed in Section~\ref{s:cg}, effectively solves a convex optimization problem. Gradient descent applied in the spin glass setting (discussed in Section~\ref{s:spin}) and stochastic gradient descent in the context of deep learning (MNIST, discussed in Section~\ref{s:mnist}) are much more complicated non-convex optimization processes. Despite the fact that these algorithms share very little geometry in common, they all (empirically) exhibit universality in halting time (Table \ref{f:CG-table}).  Indeed, if the normalized third and fourth moments are close, there is a strong indication that fluctuations are universal.  

\begin{table}[h]
  \begin{center}
  \begin{small}
  \begin{sc}
    \begin{tabular}{lccccc}
      Model & Ensemble/Input Distribution & Mean & St.dev. & 3rd & 4th\\
      \hline
      CG: $M = N$ & LOE & 970 & 164 & 5.1 & 35.2\\
      CG: $M = N$ & LUE & 921 & 46 & 15.7 & 288.5 \\
      CG: $M = N + 2  \lfloor \sqrt N \rfloor$ & LOE & 366 & 13 & 0.08& 3.1\\
      CG: $M = N + 2  \lfloor \sqrt N \rfloor$ & LUE & 367 & 9 & 0.07& 3.0 \\
      CG: $M = N + 2  \lfloor \sqrt N \rfloor$ & PBE & 365 & 13 & 0.08& 3.0\\
      Spin Glass & Gaussian & 192 & 79.7 & 1.10 & 4.58 \\
      Spin Glass & Bernoulli & 192 & 80.2 & 1.10 & 4.56 \\
      Spin Glass & Uniform & 193 & 79.6 & 1.10 & 4.54 \\ 
      Fully connected & MNIST & 2929 & 106 & -0.32 & 3.24 \\
      Fully connected & Random & 4223 & 53 & -0.08 & 2.98\\
      Convnet & MNIST & 2096 & 166 & -0.11 & 3.18\\ 
      Cond. on gradient & MNIST & 3371 & 118 & -0.34 & 3.31\\
    \end{tabular}
  \end{sc}
  \caption{\label{f:CG-table} Skewness (normalized 3rd moment) and kurtosis (normalized 4th moment) for the halting times in the experiments performed below: (Rows 1-5, Section~\ref{s:cg}) In the $M = N + 2 \lfloor \sqrt N \rfloor$  it is clear that these normalized moments nearly coincide and they are quite distinct for $M = N$. (Rows 6-8, Section~\ref{s:spin}) The Gumbel like distribution in spin glasses. (Rows 9-12, Section~\ref{s:mnist}) Gaussian-like distribution, with a flat left tail for deep learning. Note that the first and second moments are zero and one since the date is normalized.} 
  \end{small}
  \end{center}
\end{table}

\subsection{The conjugate gradient algorithm}\label{s:cg}

The conjugate gradient algorithm \cite{hestenes1952methods} for solving the $N \times N$ linear system $Ax=b$, when $A = A^*$ is positive definite, is an iterative procedure to find the minimum of the convex quadratic form:
\begin{align*}
F(A,y) = \frac{1}{2}y^*Ay - y^*b,
\end{align*}
where $^*$ denotes the conjugate-transpose operation. Given an initial guess $x_0$ (we use $x_0 = b$), compute $r_0 = b - Ax_0$ and set $p_0 = r_0$. For $k = 1, \ldots, N$,
\begin{enumerate}
\item Compute $r_k = r_{k-1} - a_{k-1} A p_{k-1}$ where $a_{k-1} = \displaystyle {\langle r_{k-1}, r_{k-1} \rangle}/{\langle p_{k-1}, A p_{k-1} \rangle}$.
\item Compute $p_k = r_k + b_{k-1} p_{k-1}$ where $b_{k-1} = \displaystyle {\langle r_k,r_k \rangle}/{\langle r_{k-1},r_{k-1} \rangle}$.
\item Compute $x_k = x_{k-1} + a_{k-1} p_{k-1}$.
\end{enumerate}


If $A$ is strictly positive definite $x_k \to x = A^{-1} b$ as $k \to \infty$. Geometrically, the iterates $x_k$ are the best approximations of $x$ over larger and larger affine Krylov subspaces $\mathcal K_k$,
\begin{align*}
  \|Ax_k-b\|_A &= \mathrm{min}_{x \in \mathcal K_k} \|Ax-b\|_A, \quad \mathcal K_k = x_0 + \mathrm{span}\{r_0,Ar_0,\ldots,A^{k-1}r_0\}, \quad \|x\|_A^2 = \langle x, A^{-1} x \rangle, 
\end{align*}
as $k \uparrow N$. The quantity one monitors over the course of the conjugate gradient algorithm is the norm  $\|r_k\|$:
\begin{align*}
  T_{\epsilon,N,\mathrm{CG},E}(A,b) := \min\{k: \|r_k\| < \epsilon\}.
\end{align*}

In exact arithmetic, the method takes at most $N$ steps: In calculations with finite-precision arithmetic the number of steps can be much larger than $N$ and the behavior of the algorithm in finite-precision arithmetic has been the focus of much research \cite{greenbaum1989behavior, greenbaum1992predicting}. What is important for us here is that it may happen that $\|r_k\| < \epsilon$ but the true residual $\hat r_k := b-A x_k$ (which typically differs from $r_k$ in finite-precision computations) satisfies $\|\hat r_k\| > \epsilon$.

\begin{figure}[tbp]
\begin{center}
  \subfigure[]{\includegraphics[width=0.45\linewidth]{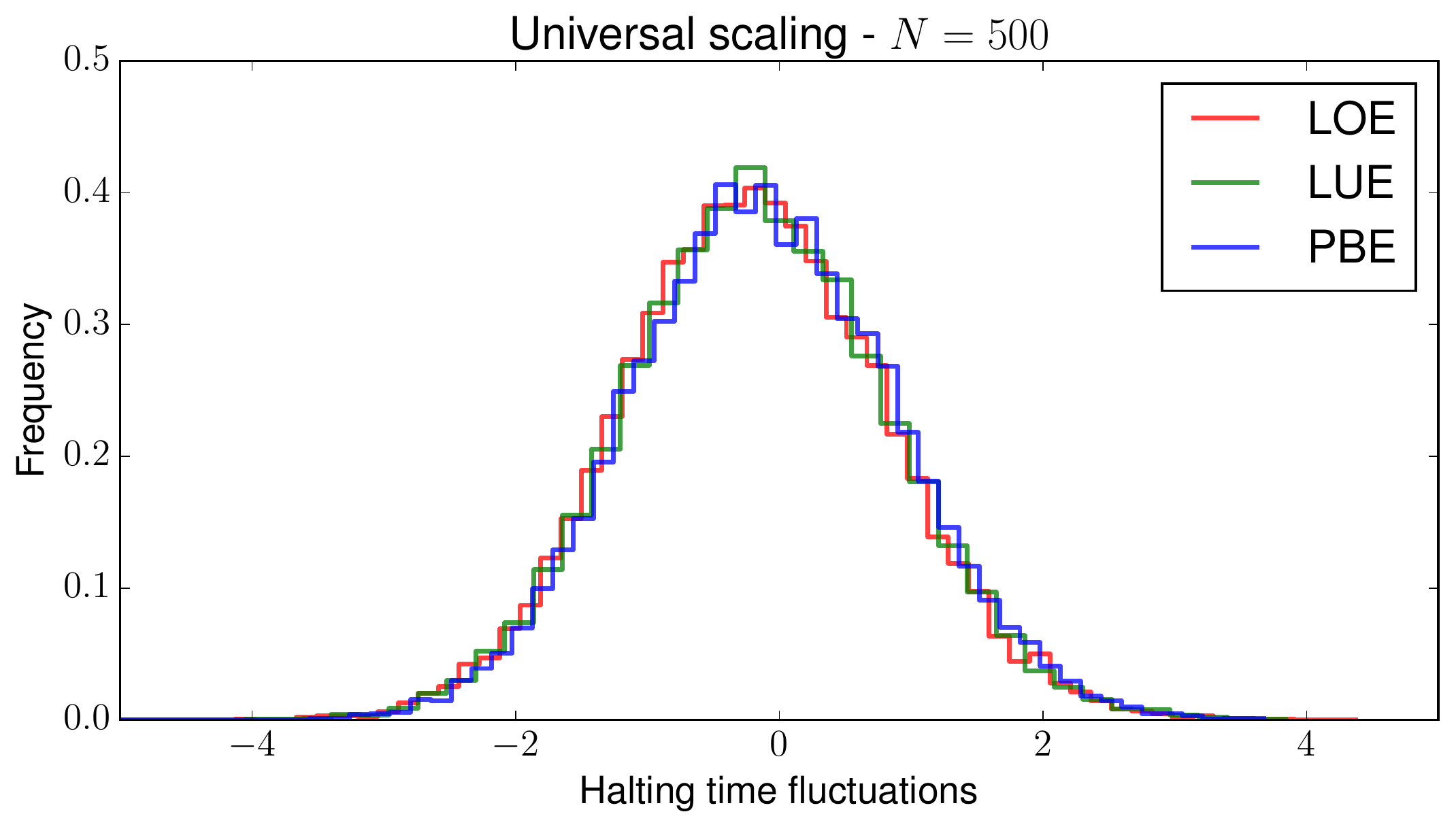} \label{f:cg-crit}}
  \subfigure[]{\includegraphics[width=0.45\linewidth]{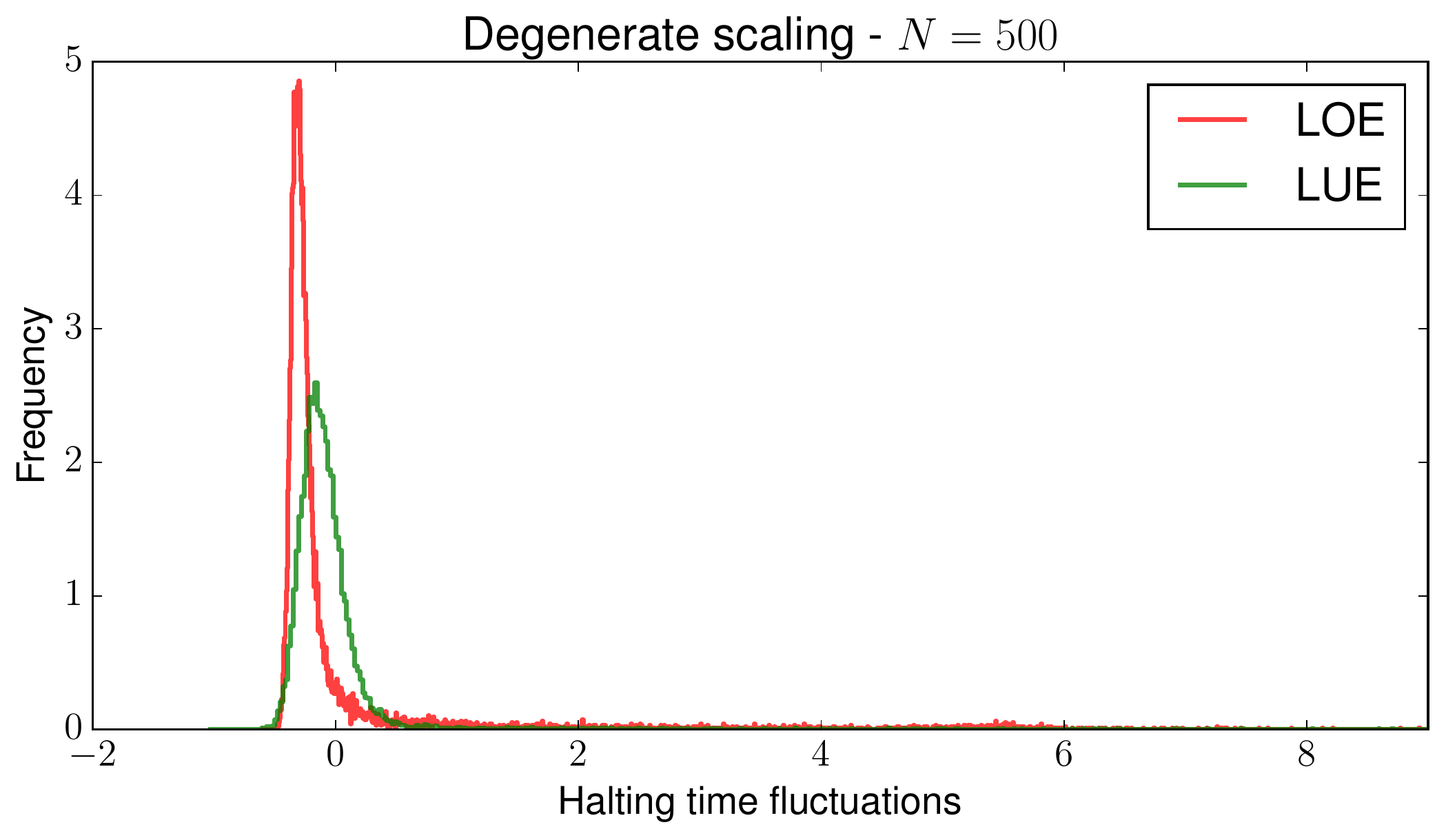} \label{f:cg-sing}}
\end{center}
\caption{Empirical histograms for the halting time fluctuations $\tau_{\epsilon,N,\mathrm{CG},E}$ when $N = 500$, $\epsilon = 10^{-10}$ for various choices of ensembles $E$.   (a) The scaling $M = N + 2 \lfloor \sqrt N \rfloor$ demonstrating the presence of universality.  This plot shows three histograms, one each for $E=$ LUE, LOE and PBE. (b) The scaling $M = N$ showing two histograms for $E=$ LUE and LOE and demonstrating the non-existence of universality.}
\end{figure}

Now, we discuss our choices for ensembles $E$ of random data.  In all computations, we take $b = (b_j)_{1\leq j \leq N}$ where each $b_j$ is iid uniform on $(-1,1)$. We construct positive definite matrices $A$ by $A = XX^*$ where $X = (X_{ij})_{1\leq i \leq N,~1\leq j \leq M}$ and each $X_{ij} \sim \mathcal D$ is iid for some distribution $\mathcal D$.  We make the following three choices for $\mathcal D$, Positive definite Bernoulli ensemble (PDE), Laguerre orthogonal ensemble (LOE), Laguerre unitary ensemble (LUE):

\begin{itemize}
\item[PBE] $\mathcal D$ a Bernoulli $\pm 1$ random variable (equal probability).
\item[LOE] $\mathcal D$ is a standard normal random variable.
\item[LUE] $\mathcal D$ is a standard complex normal random variable.
\end{itemize}

The choice of the integer $M$, which is the inner dimension of the matrices in the product $XX^*$, is critical for the existence of universality. In \cite{deift2014universality} and \cite{deift2015condition} it is demonstrated that universality is present when $M = N + \lfloor c \sqrt N \rfloor$ and $\epsilon$ is small, but fixed. Universality is not present when $M = N$ and this can be explained heuristically by examining the distribution of the condition number of the matrix $A$ in the LUE setting \cite{deift2015condition}.  We demonstrate this again in Figure~\ref{f:cg-crit}.  We also demonstrate that universality does indeed fail\footnote{For those familiar with random matrix theory, this might not be surprising as real and complex matrices typically lie in different universality classes.  From this point of view, it is yet more striking that Figure~\ref{f:cg-crit} gives a universal curve.} for $M = N$ in Figure~\ref{f:cg-sing}.  We refer to Table~\ref{f:CG-table}(Rows 1-5) for a quantitative verification of universality.

\subsection{Spin glasses and gradient descent}\label{s:spin}

The gradient descent algorithm for the Hamiltonian of the $p$-spin spherical glass will find a local minimum of the non-convex function \eqref{ham}. Since variance the of $H_{N}(w)$ is typically of order $N$, a local minimum scales like $-N$. More precisely, from \cite{auffinger2013random}, the energy of the floor level where most of local minima are located is asymptotically at $-2\sqrt{2/3}N\approx-1.633N$ and the ground state is around $-1.657N$. The algorithm starts by picking a random element $w$ of the sphere with radius $\sqrt{N}$, $S^{N-1}(\sqrt{N})$, as a starting point for each trial. We vary the environment for each trial and introduce ensembles by setting $x_{(\LargerCdot)} \sim \mathcal D$ for a number of choices of distributions $\mathcal D$. For a fixed dimension $N$, accuracy $\epsilon$ that bounds the norm of the gradient, and an ensemble $E$: (1)  Calculate the gradient steps: $w^{t+1} = w^t - \eta_t \nabla_wH(w^t)$, (2) Normalize the resulting vector to the sphere: $\sqrt{N}\frac{w^{t+1}}{||w^{t+1}||} \leftarrow w^{t+1}$, and (3) Stop when the norm of the gradient size is below $\epsilon$ and record $T_{\epsilon,N,\mathrm{GD},E}$. This procedure is repeated 10,000 times for different ensembles (i.e. different choices for $\mathcal D$). Figure~\ref{f:SG-univ} exhibits the universal halting time which presents evidence that $\tau_{\epsilon,N,\mathrm{GD},E}$ is independent of the ensemble.  Again, we refer to Table~\ref{f:CG-table}(Rows 6-8) for a clear quantitative verification of universality.

\subsection{Digit inputs vs. random inputs in deep learning}\label{s:mnist}

\begin{figure}[tbp]
\begin{center}
\subfigure[]{\includegraphics[width=0.49\linewidth]{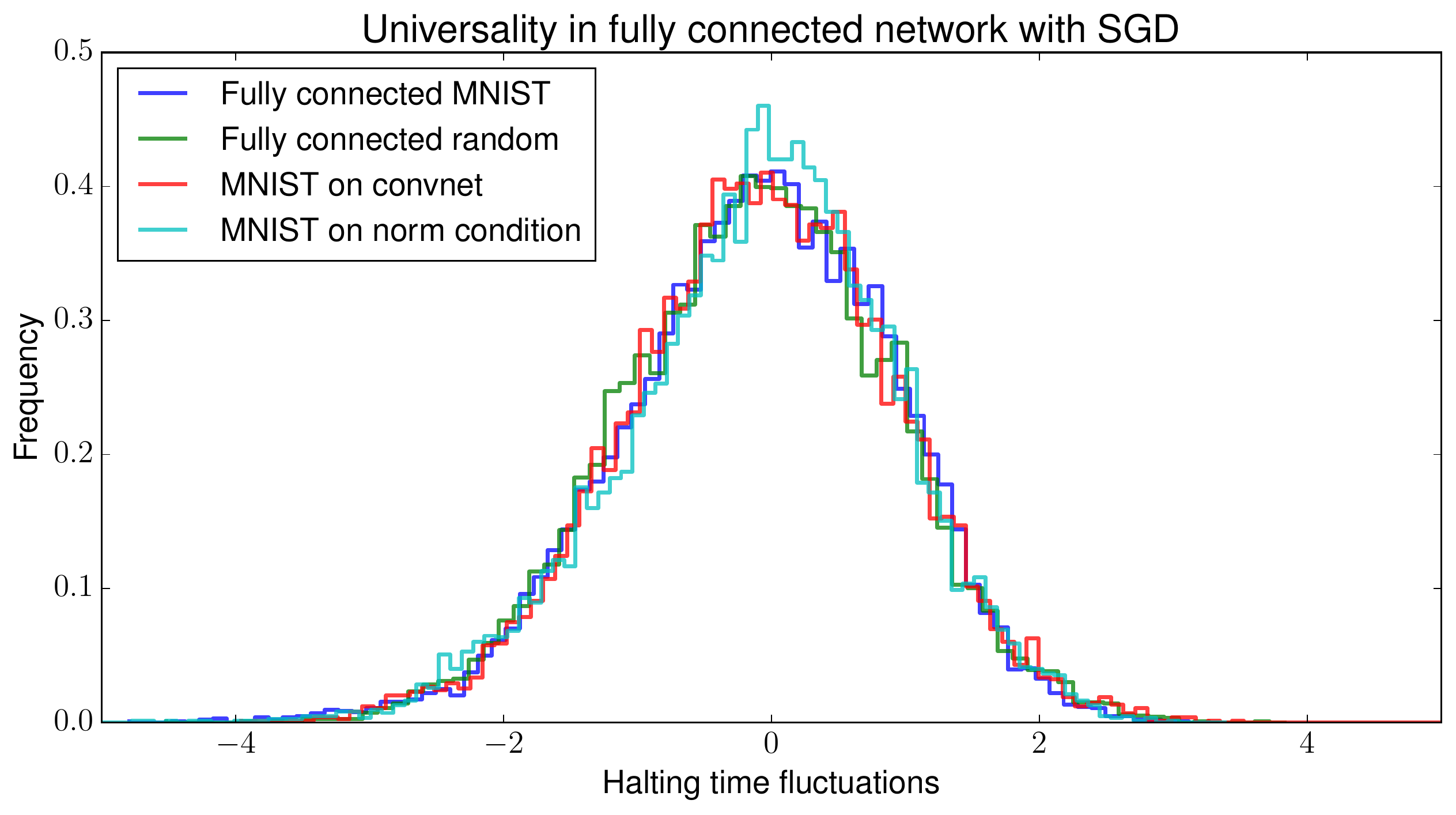}\label{f:SG-univ}}
\subfigure[]{\includegraphics[width=0.49\linewidth]{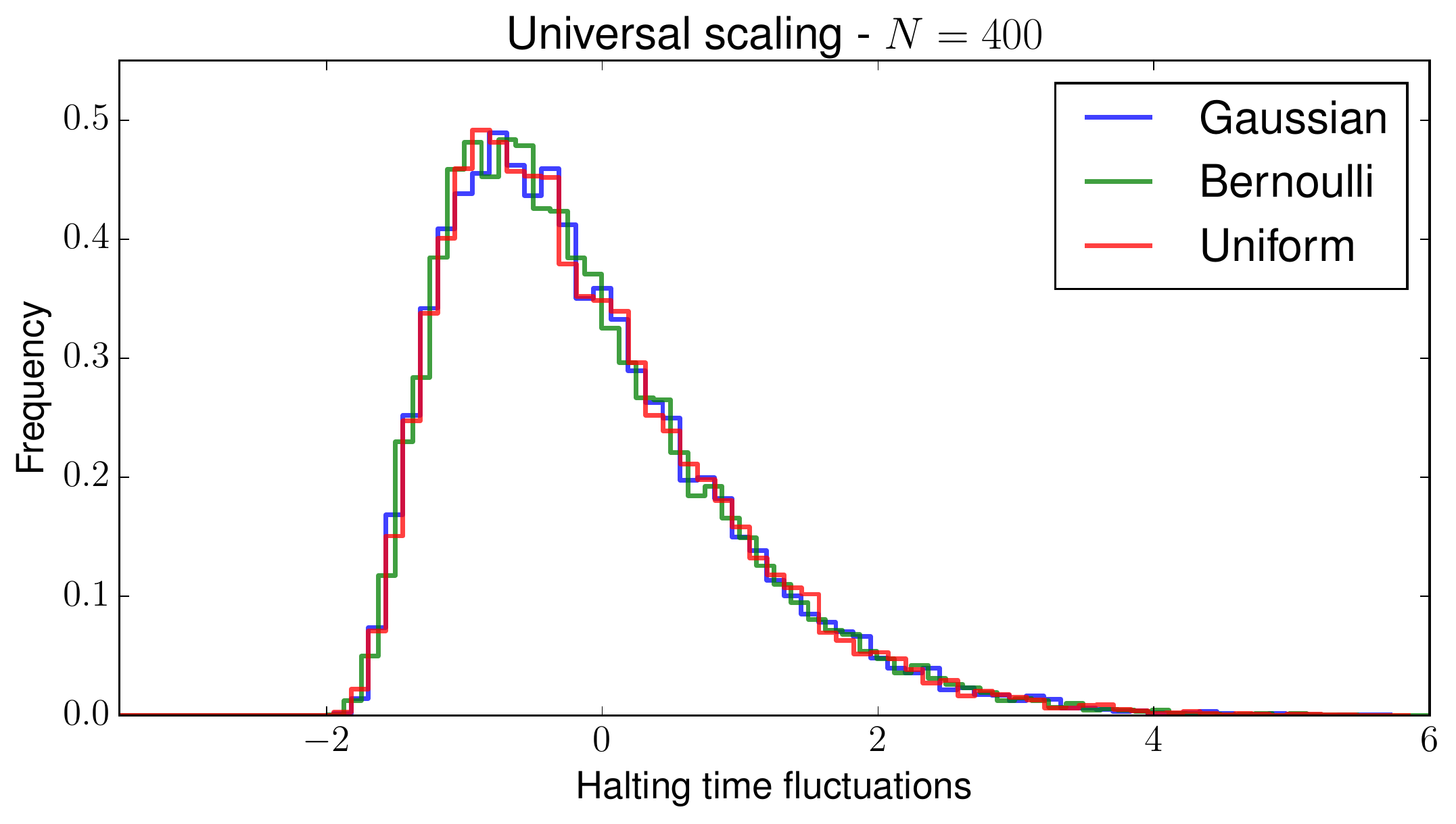}\label{f:MNIST-univ}}
\end{center}
\caption{ (a) Universality across different distributions: We choose $\mathcal D \sim \text{Gaussian}(0,1)$, $\mathcal D \sim $ uniform on $(-(3/2)^{1/3},(3/2)^{1/3})$ and $\mathcal D \sim$ Bernoulli $\pm 1/\sqrt{2}$ with equal probability. (b) Universality in the halting time for deep learning cost functions. MNIST digit inputs and independent Gaussian noise inputs give rise to the same halting time fluctuations, as well as a convnet with a different stopping condition.}
\end{figure}

A deep learning cost function is trained on two drastically different ensembles. The first is the MNIST dataset, which consists of 60,000 samples of training examples and 10,000 samples of test examples. The model is a fully connected network with two hidden layers, that have 500 and 300 units respectively. Each hidden unit has rectified linear activation, and a cross entropy cost is attached at the end. To randomize the input data we sample 30,000 samples from the training set each time we set up the model and initialize the weights randomly. Then we train the model by the stochastic gradient descent method with a minibatch size of 100. This model gets us about 97\% accuracy without any further tuning. The second ensemble uses the same model and outputs, but the input data is changed from characters to independent Gaussian noise. This model, as expected, gets us only about 10\% accuracy: it randomly picks a number! The stopping condition is reached when the average of successive differences in cost values goes below a prescribed value. As a comparison we have also added a deep convolutional network (convnet), and we used the fully connected model with a different stopping condition: one that is tied to the norm of the gradient. Figure~\ref{f:MNIST-univ} demonstrates universal fluctuations in the halting time in all of the four cases. Again, we refer to Table~\ref{f:CG-table}(Rows 9-12) for a quantitative verification of universality --- despite the large amount of noise in the dataset the skewness and kurtosis remain very close across different ensembles.  

\subsubsection*{Acknowledgments}
We thank Percy Deift for valuable discussions and G\'erard Ben Arous for his mentorship throughout the process of this research. The first author thanks very much to U\u{g}ur G\"uney for his availability for support and valuable contributions in countless implementation issues. This work was partially supported by the National Science Foundation under grant number DMS-1303018 (TT).

\bibliography{universality}
\bibliographystyle{plain}

\end{document}